\documentclass{article}

\usepackage[preprint]{neurips_2020}

\usepackage[utf8]{inputenc} %
\usepackage[T1]{fontenc}    %
\usepackage{hyperref}       %
\hypersetup{
   breaklinks=false,   %
   colorlinks=true,   %
   pdfusetitle=true,  %
}

\usepackage{url}            %
\usepackage{booktabs}       %
\usepackage{amsfonts}       %
\usepackage{nicefrac}       %
\usepackage{microtype}      %

\title{Exploring the Role of the Bottleneck in Slot-Based Models Through Covariance Regularization}

\newcommand*\samethanks[1][\value{footnote}]{\footnotemark[#1]}

\author{
    Andrew Stange\thanks{~~Equal contribution.} \quad
    Robert Lo\samethanks \quad
    Abishek Sridhar\thanks{~~Equal contribution.} \quad
    Kousik Rajesh\samethanks \quad
    \\
    School of Computer Science, Carnegie Mellon University\quad\quad \\
  {\tt \{astange,chifanl,abisheks,kousikr\}@cs.cmu.edu} \\
}

\usepackage{todonotes}
\usepackage{bbm}
\usepackage{booktabs}
\usepackage{float}

\usepackage{xspace}
\newcommand{\covloss}{\textsc{covLoss}\xspace}
\newcommand{\cosineLoss}{\textsc{cosineLoss}\xspace}
\newcommand{\cmdsprite}{\textsc{coloredMdSprites}\xspace}
\newcommand{\coco}{\textsc{coco}\xspace}
\newcommand{\fgari}{\textsc{FG-ARI}\xspace}
\newcommand{\baseline}{\textsc{Slot-Attention}\xspace}
\newcommand{\dinosaur}{\textsc{DINOSAUR}\xspace}

\begin{document}

\maketitle

\begin{abstract}

    In this project we attempt to make slot-based models with an image reconstruction objective competitive with those that use a feature reconstruction objective on real world datasets. We propose a loss-based approach to constricting the bottleneck of slot-based models, allowing larger-capacity encoder networks to be used with Slot Attention without producing degenerate stripe-shaped masks. 
    We find that our proposed method offers an improvement over the baseline Slot Attention model but does not reach the performance of \dinosaur on the COCO2017 dataset. Throughout this project, we confirm the superiority of a feature reconstruction objective over an image reconstruction objective and explore the role of the architectural bottleneck in slot-based models.
    \footnote{We make the code for this project available through the following link: 
    \url{https://github.com/robert1003/slot-attention-disentanglement}
    }

\end{abstract}

\section{Introduction}
\label{sec:intro}

Object-centric representations hold the potential to significantly improve the generalization capabilities of computer vision models through their ability to factorize and represent a scene as the composition of objects. \cite{bindingProblem} \cite{higherCog} \cite{modelbasedRL} Despite growing interest \cite{slotAttention} \cite{slate} \cite{steve} \cite{savi} \cite{saviPlusPlus}, most slot-based models still struggle on scenes with complex textures and on real world images. \cite{clevrtex} \cite{IBComplexTexture} Recent work has shown that strong encoders are likely needed to produce latent features that are clusterable by Slot Attention from these complex scenes. \cite{dinosaur} \cite{invariantSA} However, these stronger encoders loosen the architecture's information bottleneck and reduce the incentive for object separation between slots. This tradeoff between a stronger encoder and object separation is an artifact of a reliance on architectural approaches to changing the strength of the bottleneck.

In this paper, we attempt to overcome this tradeoff through the use of a loss function-based approach to constricting the bottleneck. Specifically, we adapt the projection head and variance and covariance losses from VICReg \cite{vicreg} to slot-based methods. By placing additional restrictions on the slot features, the reconstruction task is made more difficult and the bottleneck is constricted.

Our original hypothesis was that our proposed loss would improve instance segmentation performance and produce slot representations that improve downstream performance. %
While we do not achieve these results in this work, we confirm the intuitions about the role of the bottleneck in slot-based architectures underlying these predictions and present some interesting preliminary results.

\section{Related Work}
\label{sec:related}

Slot Attention \cite{slotAttention} is a convolutional autoencoder that iteratively applies a modified attention mechanism to latent vectors in order to obtain a permutation-invariant set of object-specific representations, called slots. The slot attention module relies on a variation of dot-product attention that treats slots as queries that compete to "explain away" the encoder output. While initial iterations of this architecture fail on datasets with complex textures \cite{clevrtex}, recent work demonstrates that a stronger backbone allows this architecture to achieve a more competitive performance. \cite{invariantSA}

\dinosaur \cite{dinosaur} is a slot-based architecture that relies on a frozen DINO ViT \cite{dino} as its encoder and trains using a DINO-feature reconstruction loss. %
The feature reconstruction objective is justified through an experiment where Slot Attention and \dinosaur with a MLP decoder are trained on COCO using a frozen ViT encoder with an image reconstruction and DINO-feature reconstruction objective, respectively. While Slot Attention learns a degenerate stripe pattern, the feature reconstruction objective learns instance segmentation. Despite its feature reconstruction objective, \dinosaur still suffers from training instabilities, preventing the end-to-end training of the architecture. \cite{dinosaur}

VICReg \cite{vicreg} is a self-supervised, non-contrastive representation learning method with a joint embedding architecture that relies only on its loss function to prevent collapse. This is a simpler architecture than other non-contrastive methods such as BYOL \cite{byol} or SimSiam \cite{simsiam} which frequently rely on momentum encoders, stop gradients, quantization, or batch normalization in order to avoid collapse.

There have been a number of attempts to incorporate contrastive losses into slot-based models for video \cite{cswm} \cite{slotcon}. %
Unlike these works, we apply a non-contrastive objective to slot representations as a sort of regularization term and are not restricted to training on video.

\section{Method}
\label{sec:method}

\begin{figure}[ht]
\centering
\includegraphics[scale=0.2]{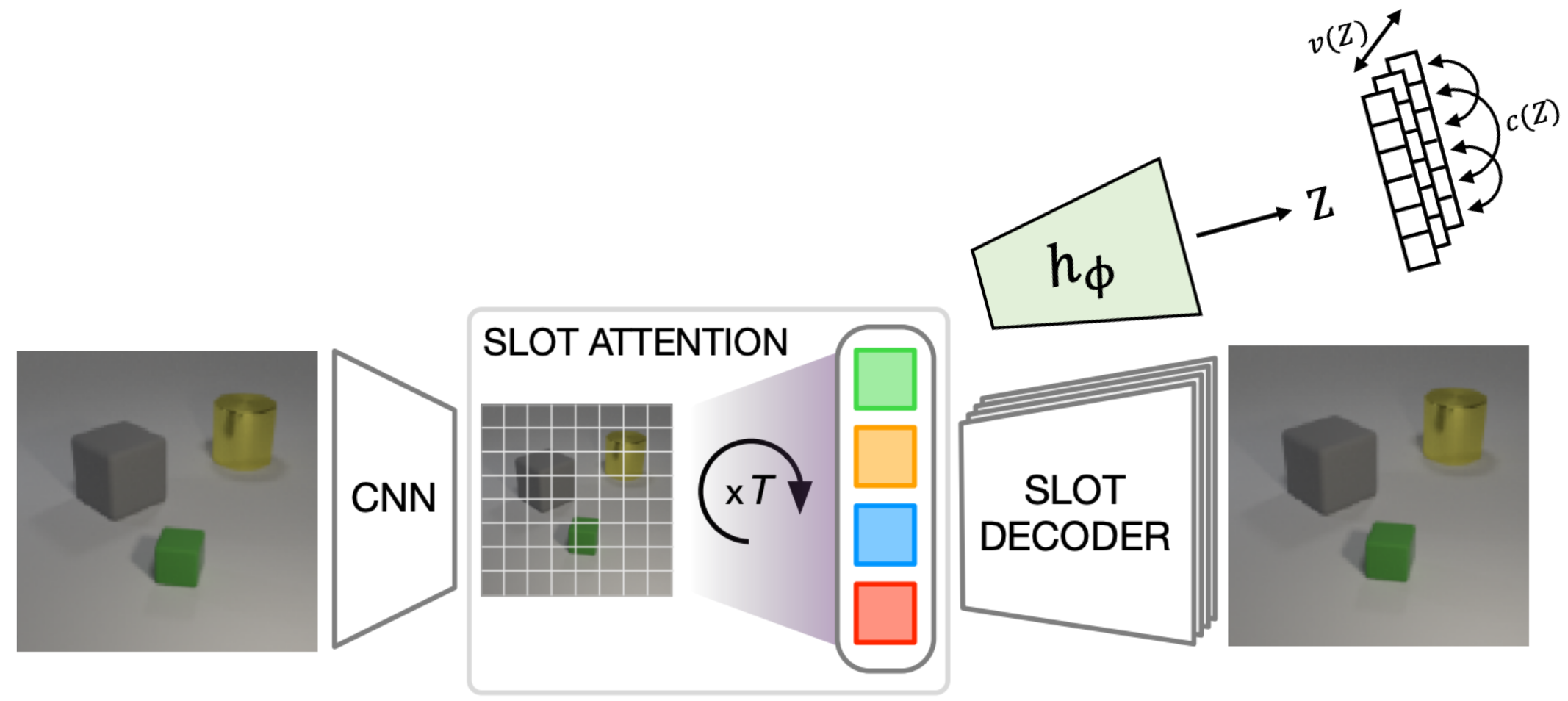}
\caption{Apply the VICReg projection head and losses to the output of the slot attention module. \cite{slotAttention}\cite{vicreg}}
\label{fig:method}
\end{figure}

Slot Attention produces a representation of an input image using a convolutional encoder and then iteratively applies the slot attention module in order to obtain a permutation-invariant set of $K$ slots, each with $D_{slots}$ dimensions. \cite{slotAttention} %
We propose the addition of a MLP projection head $h_{\phi}:\mathbb{R}^{D_{slots}} \to \mathbb{R}^{D_{proj}}$ where $D_{proj} \gg D_{slots}$ (see Figure \ref{fig:method}). In this projection space, we will compute the variance and covariance losses:
\[
v(Z) = \frac{1}{D_{proj}} \sum_{j=1}^{D_{proj}} \max\left(0, \gamma - \sqrt{Var(z_j) + \epsilon}\right)
\]
$$c(Z) = \frac{1}{D_{proj}} \sum_{i \neq j} [Cov(Z)]^2_{i, j}$$

where $z_j$ is the $j^{th}$ dimension of the projection space, $\gamma$ is a hyperparameter representing the desired variance for each feature, and $\epsilon$ is a small scalar included for numerical stability. Covariance is calculated over the features of the projection space, giving a covariance matrix of size $D_{proj} \times D_{proj}$. The total loss function for our method is: 

$$\mathcal{L}(X, \hat{X}) = ||X - \hat{X}||^2_2  + \beta\left(v(Z) + c(Z)\right)$$

where $X$ is an input image, $\hat{X}$ is the reconstruction, $Z$ is the projection of the slot representations for $X$ through $h_{\phi}$, and $\beta$ is a hyperparameter that weights the variance and covariance losses and dictates the strength of the information bottleneck.

Empirically we find that the variance loss term is not necessary for our task. While VICReg \cite{vicreg} finds that the variance loss term helps avoid representation collapse, we find that our primary reconstruction objective is sufficient to prevent collapse and that the variance loss can be removed. We call this loss using only the reconstruction and covariance terms \covloss.

In addition to \covloss, we test another loss called \cosineLoss:
$$
cosine(Z) = \sum_{i=1}^{D_{slots}}\sum_{j=1}^{D_{slots}} \mathbbm{1}[i \neq j] cosemb(z_i, z_j)
$$
where $cosemb(z_i, z_j) = \max(0, \cos(z_i, z_j)-0.2)$. This loss is similar to the InfoNCE loss \cite{infonce}, and the intuition is to encourage the model to make embeddings of slots far from each other. Note that unlike \covloss, this loss is applied \textbf{directly to the slot feature vectors}. Attempts to apply this loss to the projection of the slot vectors diverges. Intuitively, since $D_{proj}$ can be quite large compared to $D_{slots}$, the projection head can learn to project similar slot feature vectors to constant orthogonal spaces, that can pose a local optima for the \cosineLoss and hinder the effect we seek to achieve.

\section{Experiments \& Results}
\label{sec:results}

\subsection{Datasets \& Metrics}

For the experiments we consider two dataset: \cmdsprite\cite{mdsprites} and \coco\cite{cocodataset}.

\begin{itemize}
    \item \cmdsprite: This dataset is chosen for evaluation since it provides ground truth features (2D position, shape, color, and scale) for each object in the scene and for its simplicity. The Slot Attention paper \cite{slotAttention} uses a version of this dataset with a grey background for all examples. Since performance is saturated on this grey-background dataset, we use a color-background version of the dataset, \cmdsprite \cite{multiobjectdatasets19}. The first $60000$ samples are used as training data and test performance is evaluated on the following $320$ samples, following the evaluation method of Slot Attention \cite{slotAttention}.
    
    \item \coco: This dataset is used to evaluate the real-world image performance of our method as the original Slot Attention method has been shown to fail miserably on it. \cite{dinosaur} We hypothesize that Slot Attention's issues with stripe-shaped masks on this dataset can be ameliorated using our method.
\end{itemize}

Following prior work \cite{slotAttention}, we use the foreground Adjusted Rand Index (\fgari) as a metric to measure the similarity of the predicted and ground truth masks. \fgari only evaluates mask overlap with foreground objects and does not evaluate the background mask. 

\subsection{\cmdsprite Experiments}

\subsubsection{Reconstruction and Mask Quality}

\begin{figure}[htbp]
  \centering
  \begin{minipage}[b]{0.9\textwidth}
  \centering
    \includegraphics[width=0.8\textwidth]{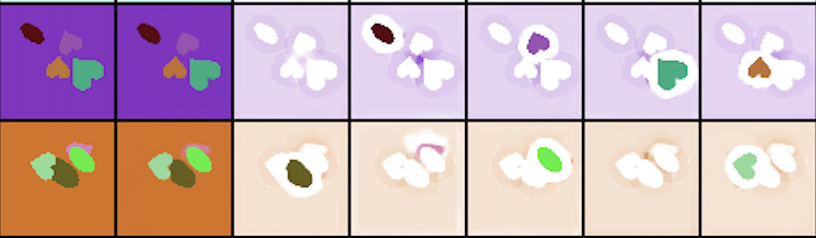}
    \caption{Baseline model reconstruction on \cmdsprite. More examples in Figure \ref{fig:cmdsprite-baseline-recon}.}
    \label{fig:cmdsprite-baseline-recon-part}
  \end{minipage}
  \begin{minipage}[b]{0.9\textwidth}
  \centering
    \includegraphics[width=0.8\textwidth]{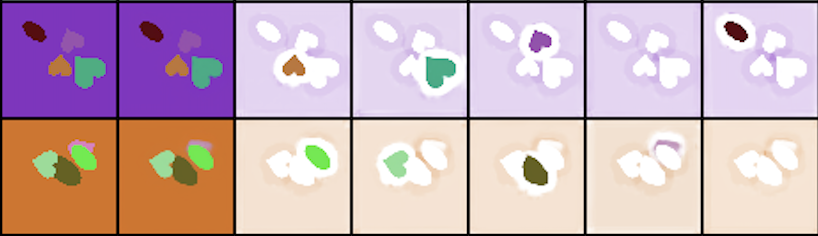}
    \caption{\covloss model reconstruction on \cmdsprite. More examples in Figure \ref{fig:cmdsprite-cov-recon}.}
    \label{fig:cmdsprite-cov-recon-part}
  \end{minipage}
  \begin{minipage}[b]{0.9\textwidth}
  \centering
    \includegraphics[width=0.8\textwidth]{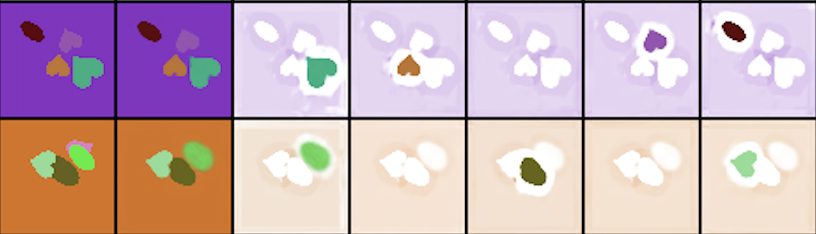}
    \caption{\cosineLoss model reconstruction on \cmdsprite. More examples in Figure \ref{fig:cmdsprite-info-recon}.}
    \label{fig:cmdsprite-info-recon-part}
  \end{minipage}
\end{figure}

Figures \ref{fig:cmdsprite-baseline-recon-part}, \ref{fig:cmdsprite-cov-recon-part}, \ref{fig:cmdsprite-info-recon-part} contain visual results of our experiments with the \cmdsprite dataset while \fgari results are shown in Table \ref{tab:ari}. Qualitatively and quantitatively, these results show that the baseline Slot Attention model outperforms both of our proposed losses on this dataset. 

While we anticipated that a tighter bottleneck would force a more efficient encoding of latent features into slot vectors, these results demonstrate that the original strength of the information bottleneck of this architecture is already sufficient to produce good object-centric representations. The bottleneck of the baseline convolutional Slot Attention architecture is already well-adjusted to this simple dataset so constricting the bottleneck further only reduces the quality of the reconstruction and the masks and hurts performance.

\subsubsection{Feature Prediction}
\label{subsubsec:feature-pred}

We evaluate the quality of the learned slot representation by training a one-hidden layer MLP to predict features of an object from its corresponding slot representation, output by a frozen slot-based model. \cite{generalizationAndRobustness} Ground truth shape, color, scale, and x and y coordinates for each object in the image are provided by the \cmdsprite dataset. The shape feature is categorical and will use accuracy as an evaluation metric. All other features are continuous and the $R^2$ value is used as an evaluation metric. We evaluate the baseline Slot Attention model, the \covloss model, and the \cosineLoss model. Since the set of slots output by a model are permutation invariant, the feature predictions from the MLP are matched to their ground truth counterparts using loss matching and the Hungarian algorithm.

Table \ref{tab:cmdsprite-feature-pred} shows the results of this experiment. \covloss is outperformed by the baseline across all features while \cosineLoss outperforms the baseline on two of the five features. Following the logic of models like $\beta$-VAE \cite{betaVAE} that produce disentangled features with a stronger bottleneck, we would have expected a tighter bottleneck to produce better features. These results suggest that the projection head used for \covloss may have too much capacity. While this projection head should have at least one nonlinearity so nonlinear correlations are penalized by the covariance loss, our current architecture with two hidden layers may have sufficient capacity to perform any transformations needed to reduce the project space covariance without meaningfully impacting the correlations between slot features.

\begin{table}[]
\centering
\begin{tabular}{@{}lll@{}}
\toprule
                       & \cmdsprite \fgari & \coco \fgari    \\ \midrule
\baseline              & 90.71   & 20     \\
\covloss    & 87.42    & 29.55 \\
\cosineLoss & 88.82    & 34.23      \\ \bottomrule
\end{tabular}
\caption{ARI on \cmdsprite and \coco}
\label{tab:ari}
\end{table}

\begin{table}[]
\centering
\begin{tabular}{@{}llllll@{}}
\toprule
                            & Shape (acc.)  & Color ($R^2$)   & Scale ($R^2$)   & X ($R^2$)   & Y ($R^2$)    \\ \midrule
\baseline                    & 0.8472        & 0.9485        & 0.7865        & 0.9734    & 0.9726      \\
\covloss                    & 0.8438        & 0.9359        & 0.7675        & 0.9641    & 0.9694 \\
\cosineLoss                 & 0.8984        & 0.9199        & 0.8124        & 0.9592    & 0.9626 \\ \bottomrule
\end{tabular}
\caption{Feature Prediction on \cmdsprite}
\label{tab:cmdsprite-feature-pred}
\end{table}

\subsection{\coco Experiments}

\subsubsection{ViT Encoder and Image Reconstruction Experiment}

\begin{figure}[htbp]
  \centering
    \includegraphics[width=\textwidth]{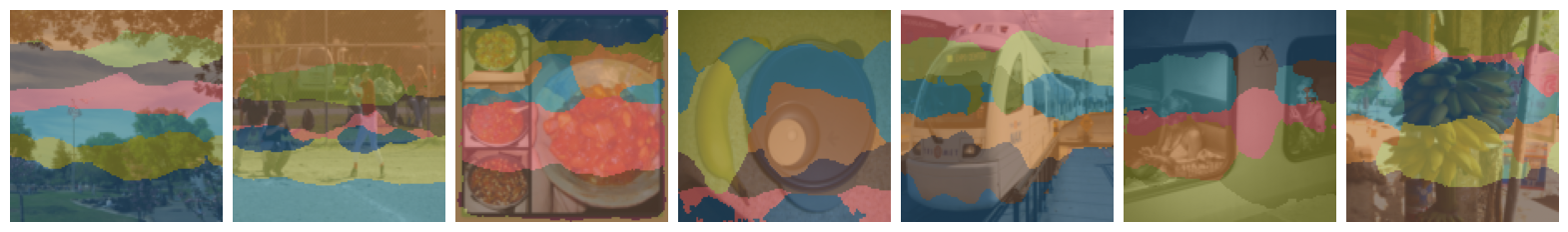}
    \includegraphics[width=\textwidth]{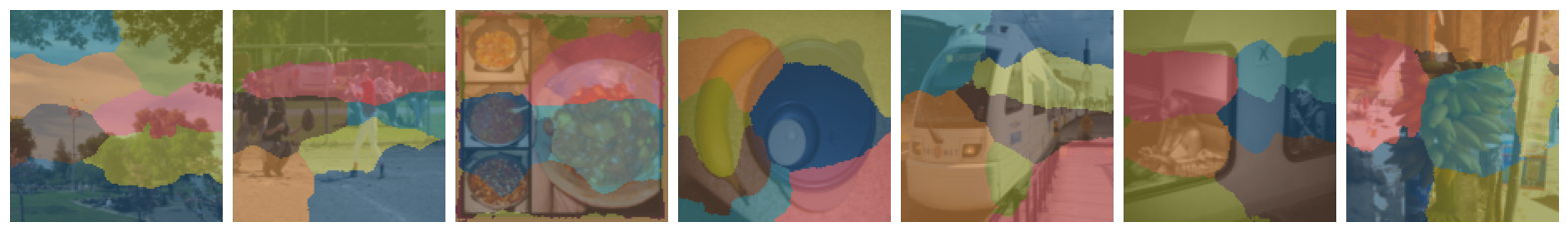}
    \includegraphics[width=\textwidth]{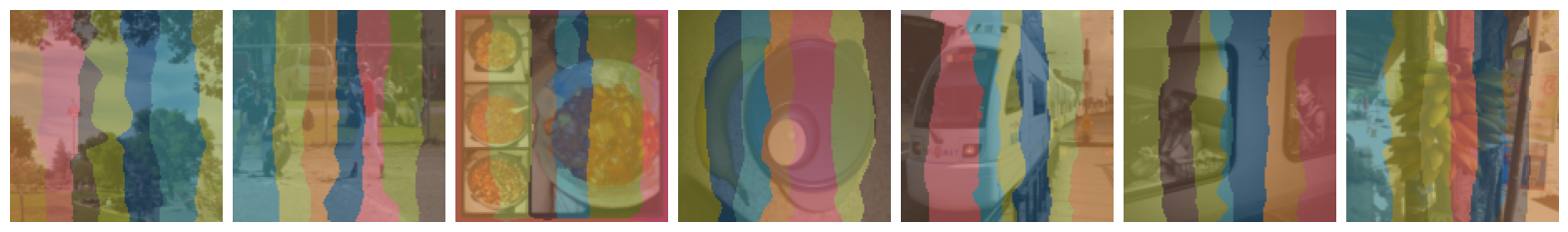}
    \includegraphics[width=\textwidth]{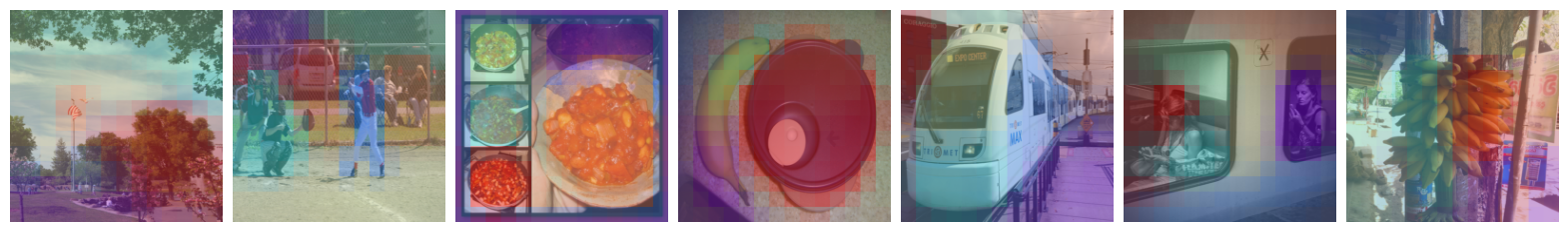}
    \includegraphics[width=\textwidth]{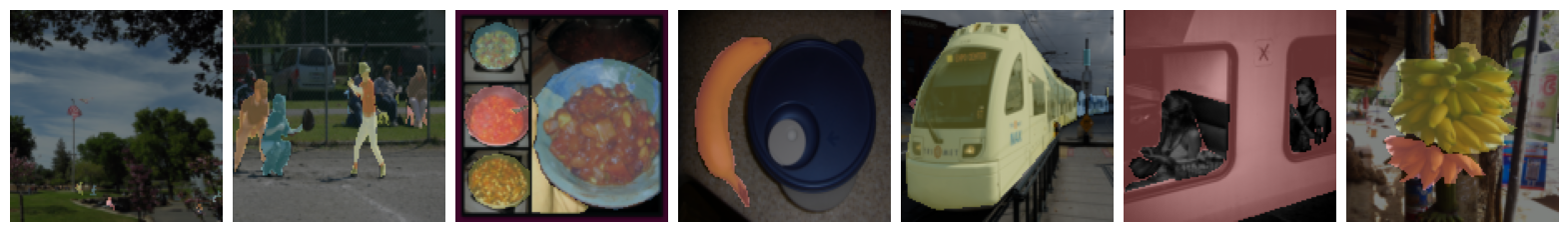}
  \label{fig:coco-example}
  \caption{\coco mask visualization for different models. Top to bottom: \baseline, \covloss, \cosineLoss, \dinosaur, ground truth. More examples in Appendix \ref{appendix:coco-mask}.}
\end{figure}

Next, we evaluate the utility of our method for scaling slot-based architectures by reproducing the \dinosaur experiment discussed in Section \ref{sec:related}. In this experiment, the convolutional encoder of Slot Attention is replaced with a frozen ViT wncoder that was pretrained with DINO. The remaining slot attention module and the decoder are then trained using an image-reconstruction objective on COCO, on top of the embeddings from this frozen encoder. \cite{dinosaur} demonstrates that Slot Attention with its image reconstruction objective fails in this setting and produces masks with a degenerate striping pattern, as seen in Figure \ref{fig:coco-example}. We hypothesize that the larger-capacity ViT encoder loosens the bottleneck of the autoencoder architecture, reducing the incentive for Slot Attention to efficiently encode the latent features into slot vectors and resulting in a failure to produce object separation between masks. For this reason, we anticipate that our additional covariance regularization loss term will be able to sufficiently constrict the bottleneck to produce object separation despite the use of a stronger ViT encoder.

To manage the computationally intensive nature of this experiment, we precompute ViT embeddings for the entire dataset a priori as the encoder is frozen. We also downsample the input images used for supervision and use a smaller hidden dimension of 64 compared to the 256 hidden dimensions used in the \dinosaur paper to make the task computationally feasible. 

The results of this experiment are available in Table \ref{tab:ari}. The baseline Slot Attention results qualitatively and quantitatively match the results given in \cite{dinosaur}.
Quantitatively, the \covloss model increases \fgari over the baseline by nearly 10 points. Notably, this also outperforms the "Block Masks" baseline given in the \dinosaur paper which simply partitions the image into a set of block-shaped masks achieves an \fgari of around 23. This indicates that our method outperforms both the Slot Attention baseline and random chance. 

From Figure \ref{fig:coco-example}, it can be seen that the masks produced by the \covloss method have some object separation but that the results are far from perfect. Although there are some caveats on these results due to our use of a smaller hidden dimension which also constricts the bottleneck, this emergence of some object separation with the use of a tighter bottleneck validates our intuitions about the role of the bottleneck in slot-based architectures when applied to real-world images.

Unfortunately, it is not clear that the \covloss method can be used to achieve much better results than are shown in this report. Even in this restricted setting with a small hidden dimension, the model is easily able to bring the covariance loss to a value on the order of 1e-10. Essentially, it is possible that our proposed loss is unable to sufficiently constrict the bottleneck to compensate for the use of the stronger encoders. As was mentioned in Section \ref{subsubsec:feature-pred}, it is possible that a weaker projection head may help alleviate this issue.

Interestingly, \cosineLoss achieves a higher \fgari than \covloss or the baseline Slot Attention model despite suffering from much worse striping than even the Slot Attention baseline. This result calls into question the utility of the \fgari metric for object discovery since the results are qualitatively much worse.  This pronounced striping may be a result of applying this loss directly on the slot features rather than in the projection space since the loss encourages the slot representations to share as little information as possible. It is unclear why this incentive results in vertically striped masks since there are plenty of objects that span multiple of these vertical masks. We would expect that the model would be encouraged to exploit these regularities in each object. Additionally, it may be the case that this \cosineLoss objective is insufficiently difficult to optimize and does not meaningfully constrict the bottleneck of the architecture.

\subsubsection{Feature Reconstruction Experiment}

Table \ref{tab:dino-ari} contains the results of applying \covloss to DINOSAUR. In other words, covariance regularization is applied to the slot representations produced by the DINOSAUR model. While \covloss slightly degrades \fgari performance of DINOSAUR when the slot dimension used in \cite{dinosaur} is used, we find that \covloss improves the training stability of DINOSAUR for large slot dimensions. With a larger slot dimension, we find that DINOSAUR collapses to a degenerate solution where all slots try to represent the entire scene and the covariance between the dimensions of the slot representations increases appreciably. This collapse results in a poor reconstruction loss and \fgari performance. When \covloss is applied to DINOSAUR in this large slot dimension setting, we find that DINOSAUR does not experience any collapse, although this larger slot dimension model still under-performs the baseline DINOSAUR model.

\begin{table}[]
\centering
\begin{tabular}{@{}lll@{}}
\toprule
                        &\coco \fgari    \\ \midrule
\dinosaur                & 43.86     \\
\dinosaur + \covloss     & 42.64 \\
\dinosaur + slot dim. 1024 & 15.25 (training unstable) \\
\dinosaur + slot dim. 1024 + \covloss   & 35.35    \\ \bottomrule
\end{tabular}
\caption{ARI on \coco for \dinosaur}
\label{tab:dino-ari}
\end{table}

\section{Future Directions and Conclusion}

The primary contribution of this project is a loss-based approach to constricting the bottleneck of slot-based models. These losses are proposed in hopes of allowing the use of the stronger encoders needed to produce cluster-able features from real-world images while still achieving object separation. While the results we show in this report are not perfect, they validate our intuitions about the role of the bottleneck in producing object separation. From our experiments, it appears that the \covloss and \cosineLoss losses are an insufficient bottleneck to produce strong object separation results, although we were unable to thoroughly explore different hyperparameter combinations and design decisions due to constraints on our computational resources. 

Other approaches to constricting the bottleneck include experimenting with different, weaker decoder architectures. Recent work like SLATE \cite{slate} proposes an alternative to the spatial broadcast decoders used throughout this project. However, these architectural approaches make it necessary to adjust the decoder architecture for each dataset or downstream task which is burdensome and can be non-trivial. Another source of future work is the exploration of objective functions besides reconstruction. While \dinosaur is a first step towards moving beyond image reconstruction objectives, it still suffers from training stability issues that prevent the entire architecture from being trained end-to-end. The intersection of slot-based models and self-supervised learning is practically unexplored and a promising direction for future research.

\label{sec:conclusion}

\newpage

\bibliography{main}
\bibliographystyle{abbrvnat}

\newpage
\appendix

\section{More Reconstructions Examples on \cmdsprite}
\label{appendix:mdsprite-recon}

\begin{figure}[H]
  \centering
    \includegraphics[width=0.6\textwidth]{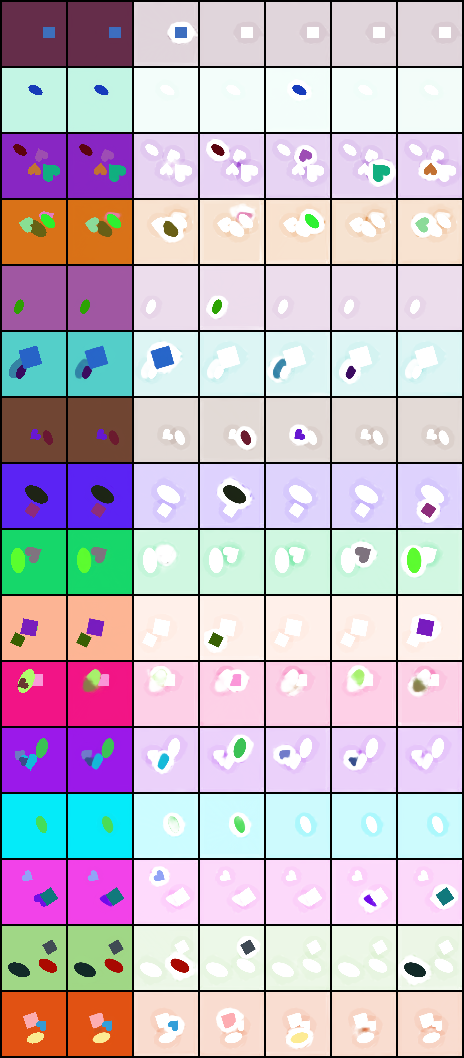}
    \caption{Baseline model reconstruction on \cmdsprite}
    \label{fig:cmdsprite-baseline-recon}
\end{figure}
\begin{figure}[H]
  \centering
    \includegraphics[width=0.6\textwidth]{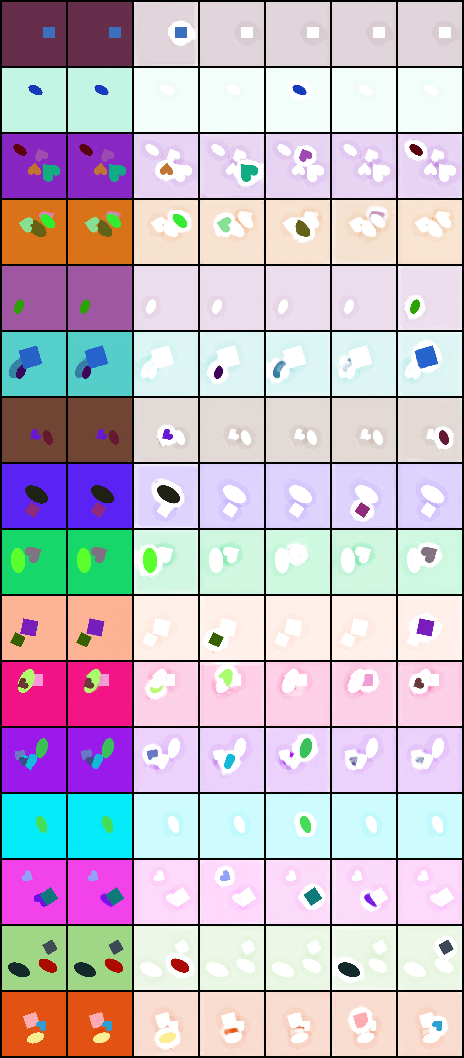}
    \caption{\covloss model reconstruction on \cmdsprite}
    \label{fig:cmdsprite-cov-recon}
\end{figure}
\begin{figure}[H]
  \centering
    \includegraphics[width=0.6\textwidth]{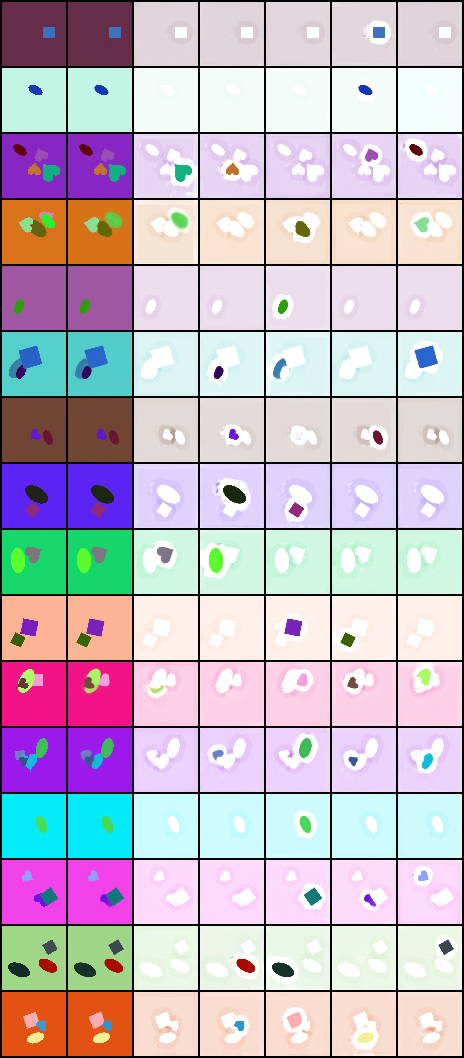}
    \caption{\cosineLoss model reconstruction on \cmdsprite}
    \label{fig:cmdsprite-info-recon}
\end{figure}

\section{More Mask Examples on \coco}
\label{appendix:coco-mask}

\begin{figure}[H]
  \centering
    Ground Truth Mask \\
    \includegraphics[width=0.77\textwidth]{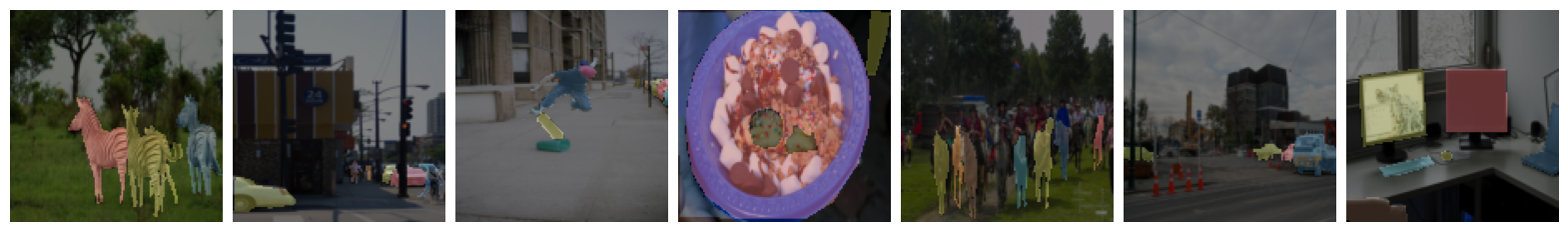}
    \includegraphics[width=0.77\textwidth]{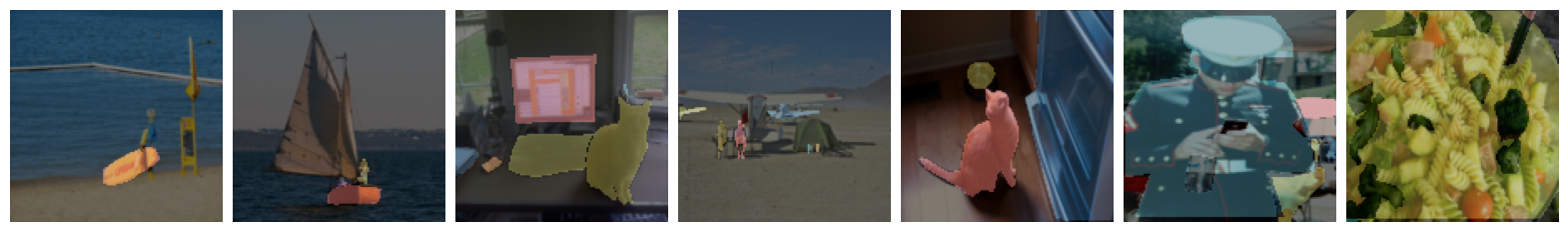}
    \includegraphics[width=0.77\textwidth]{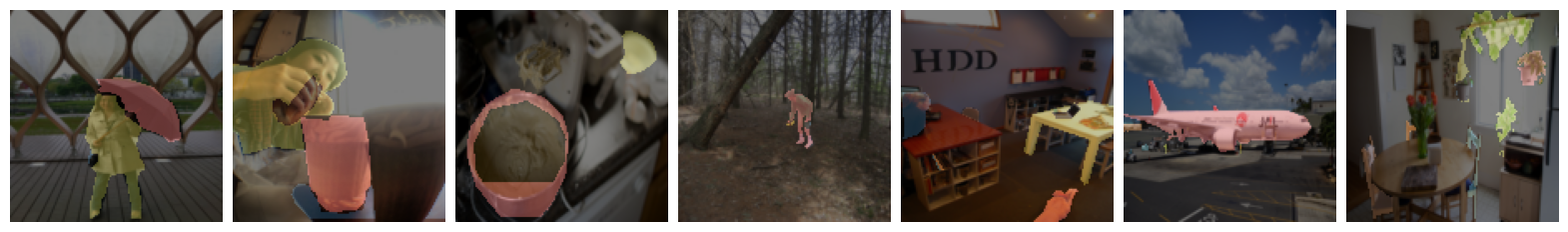} \\
    \covloss Mask \\
    \includegraphics[width=0.77\textwidth]{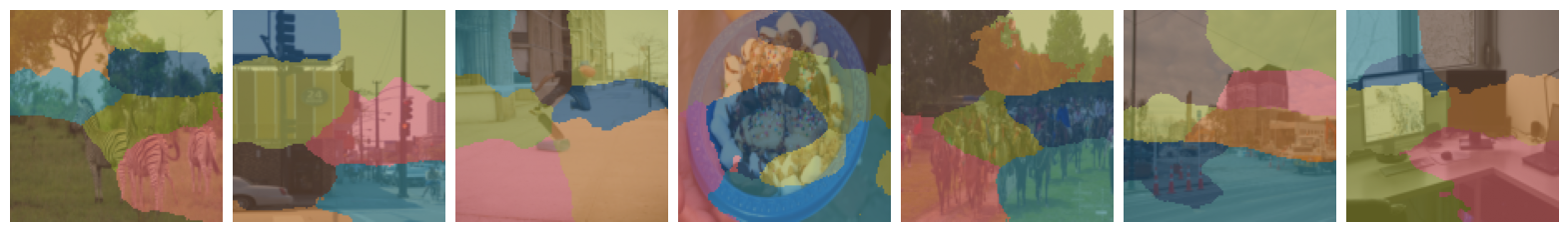}
    \includegraphics[width=0.77\textwidth]{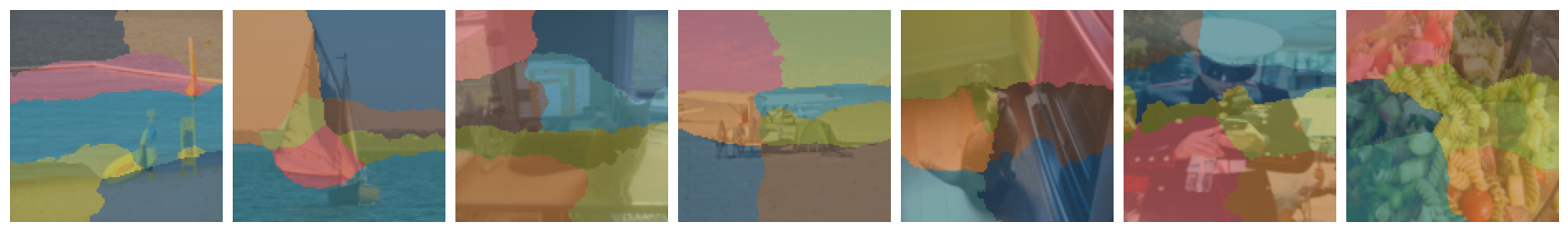}
    \includegraphics[width=0.77\textwidth]{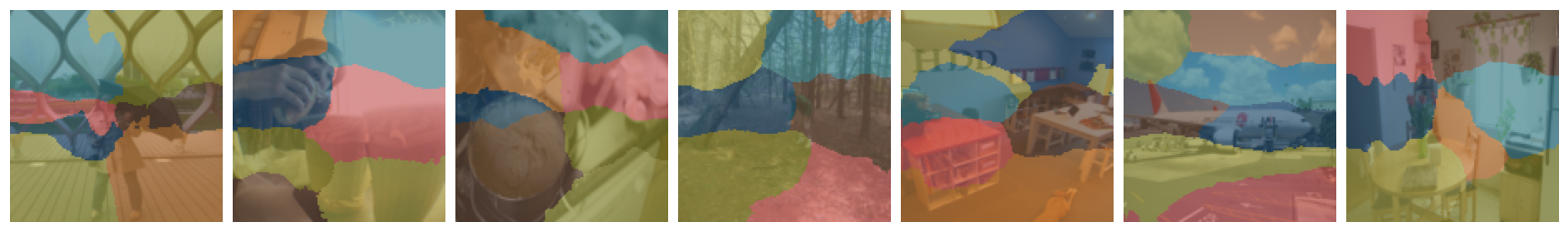} \\
    \cosineLoss Mask \\
    \includegraphics[width=0.77\textwidth]{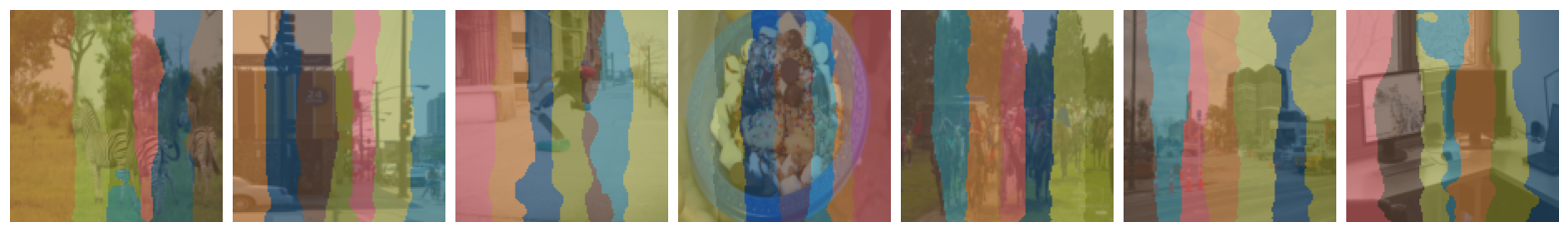}
    \includegraphics[width=0.77\textwidth]{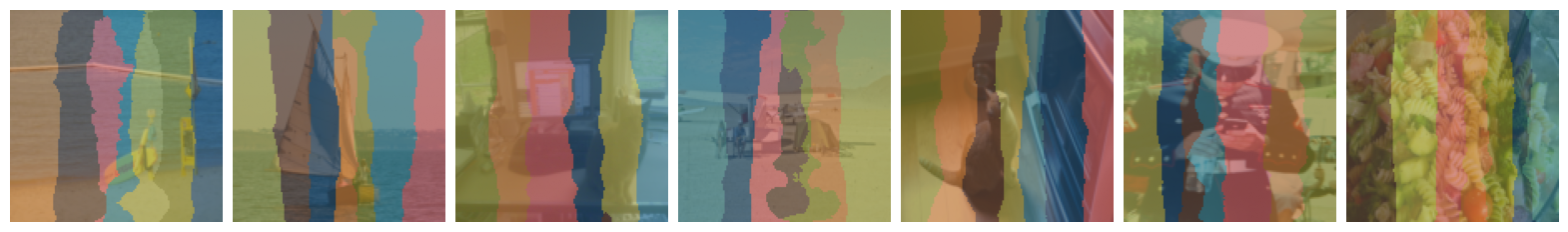}
    \includegraphics[width=0.77\textwidth]{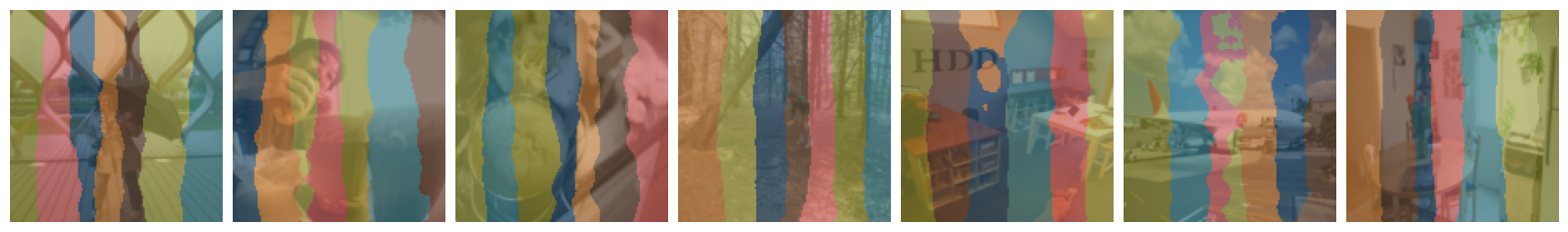} \\
    \baseline Mask \\
    \includegraphics[width=0.77\textwidth]{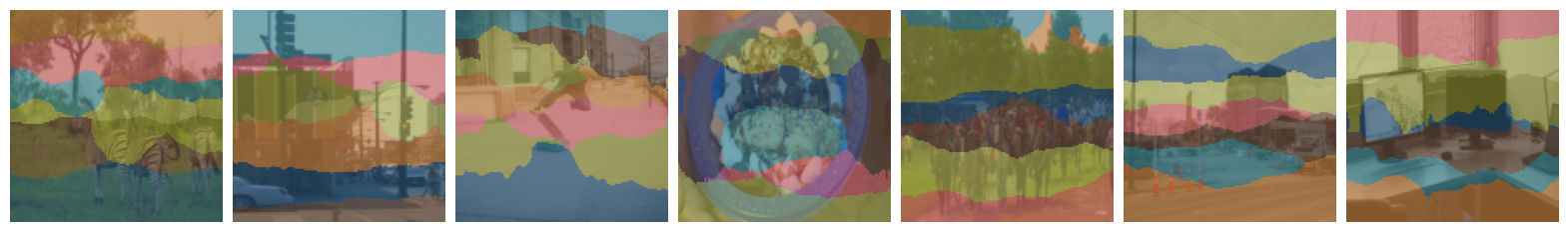}
    \includegraphics[width=0.77\textwidth]{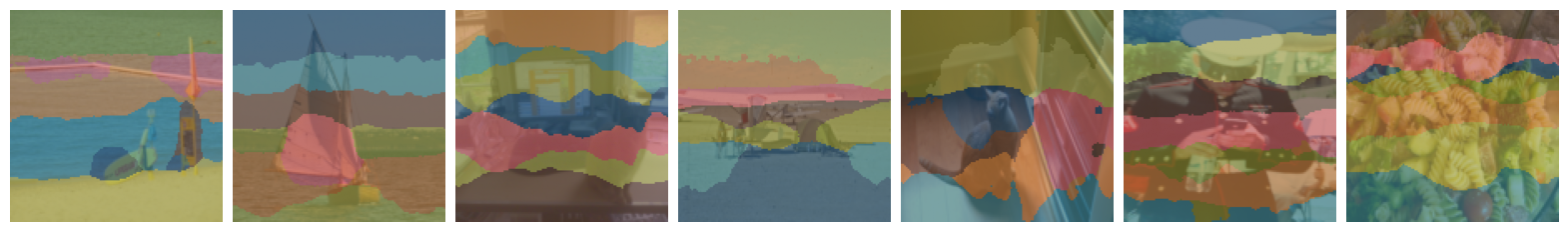}
    \includegraphics[width=0.77\textwidth]{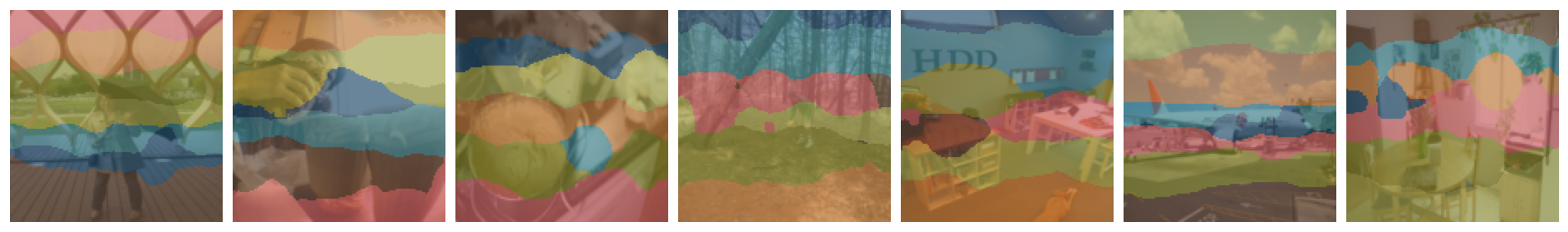} \\

    \caption{Different mask prediction on \coco}
    \label{fig:coco-mask}
\end{figure}

\begin{figure}
    \centering
    \dinosaur Mask \\
    \includegraphics[width=0.77\textwidth]{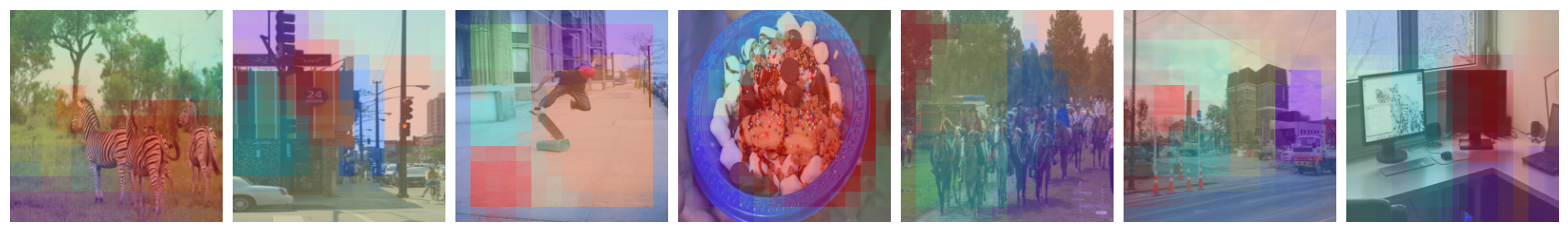}
    \includegraphics[width=0.77\textwidth]{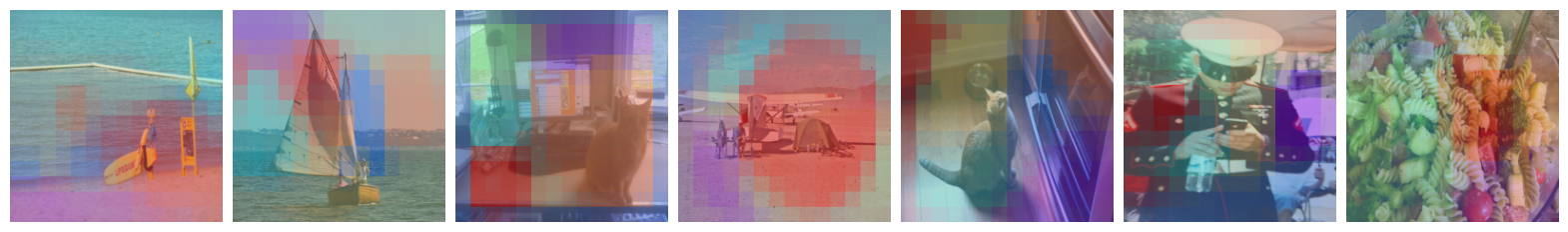}
    \includegraphics[width=0.77\textwidth]{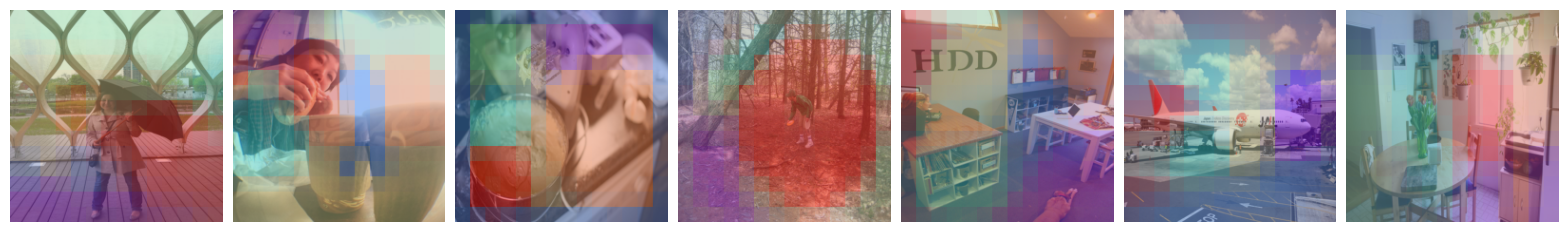} \\
    \caption{\dinosaur mask prediction on \coco}
    \label{fig:dino-coco-mask}
\end{figure}

\end{document}